\documentclass{article} 
\usepackage{iclr2027_arxiv,times}

\usepackage{amsmath,amsfonts,bm}

\def\eqref#1{equation~\ref{#1}}

\def\1{\bm{1}}

\DeclareMathAlphabet{\mathsfit}{\encodingdefault}{\sfdefault}{m}{sl}
\SetMathAlphabet{\mathsfit}{bold}{\encodingdefault}{\sfdefault}{bx}{n}

\usepackage{hyperref}
\usepackage{url}
\usepackage{graphicx}
\usepackage{multirow}
\usepackage{booktabs}

\title{Counterfactual Route Optimization for Gaussian Head Avatar Modeling}

\author{
Shikun Zhang$^{1}$,
Yong Li$^{2}$,
Yiqun Wang$^{2}$,
Cunjian Chen$^{1,*}$ \\
$^{1}$Department of Data Science and AI, Monash University, Australia \\
$^{2}$College of Computer Science, Chongqing University, China \\
$^{*}$Corresponding author
}

\iclrfinalcopy 
\begin{document}

\maketitle

\begin{abstract}
Head avatar modeling requires jointly optimizing multiple objectives with
different dominant effects on geometry, appearance, and cross-view
consistency. However, their relative effectiveness varies across training
states, while existing pipelines typically rely on fixed loss weights or
handcrafted stage-wise schedules. A central challenge is therefore to
identify which optimization direction is more beneficial at each training
state. We propose a counterfactual route optimization framework for
Gaussian head avatar modeling, which characterizes state-dependent
optimization preference from the realized effects of alternative updates
rather than predefined heuristic weighting. Starting from the same
training state, we perform short-horizon route-restricted lookahead over
geometry, appearance, and joint update routes and evaluate their outcomes
under a unified utility. The resulting counterfactual evidence is
factorized into a geometry--appearance preference and a residual joint
advantage, separately capturing the relative preference between
individual update directions and the additional benefit of coordinated
optimization. We further amortize this offline evidence into a lightweight
controller that directly estimates the current optimization preference and
applies bounded modulation to the training objectives during full avatar
optimization. Experiments on the NeRSemble dataset validate the effectiveness of the proposed design, consistently outperforming existing methods while
preserving clearer local facial structures and finer details.
\end{abstract}

\section{Introduction}

Animatable head avatars aim to faithfully reproduce identity-specific appearance and fine-grained facial dynamics while supporting controllable rendering under novel expressions, poses, and viewpoints. Such capability is important for applications including teleconferencing, immersive communication, digital entertainment, and AR/VR \citep{grassal2022neuralheadavatar, zheng2023pointavatar}. Recent advances in neural rendering have significantly broadened the representation space for head avatar modeling, progressing from implicit neural fields \citep{zheng2022imavatar} to explicit point- and Gaussian-based representations \citep{zheng2023pointavatar, qian2024gaussianavatars, xu2024gaussian}, and more recently to hybrid mesh--Gaussian formulations \citep{wang2025mega, cai2025hera}. These developments have substantially improved rendering fidelity, animation controllability, and representation efficiency. However, as avatar representations become increasingly expressive, reconstruction quality depends not only on the representation itself, but also on how the underlying optimization process is organized.

Modern head avatar training typically involves multiple objectives with different dominant effects on geometry, appearance, fine-scale details, and cross-view consistency. Importantly, the relative benefit of these optimization directions is not constant throughout training: different states may favor different update directions, while supervision that is effective at one stage may become less beneficial as the reconstruction evolves. Existing avatar methods often organize such objectives through manually designed stage-wise schedules or coarse-to-fine curricula, with their relative emphasis largely predetermined before training \citep{zhang2026fresco, ren2026omgavatar}. Although these strategies provide useful global priors and have proven effective for stabilizing avatar reconstruction, they cannot explicitly adapt the optimization preference to the actual needs of each individual training state. A central challenge is therefore to determine which optimization direction is most beneficial for the current state of avatar training.

Adaptive optimization provides a natural direction for addressing this
problem. Adaptive optimization has remained an active topic in multi-task
learning and related multi-objective settings. Beyond early
uncertainty-based weighting and gradient balancing
\citep{kendall2018multitask,chen2018gradnorm}, recent methods have
continued to estimate task importance from loss scales, gradient
characteristics, parameter-level contributions, or multi-objective
optimization criteria
\citep{thakur2025adaptive,ji2025adaptive,lin2026dual,wu2026multiobjective}.
These methods demonstrate that optimization priorities can be adjusted
dynamically rather than being entirely prescribed by hand. However, they
still typically infer the importance of different objectives from
indirect statistics or optimization signals, such as current losses,
gradients, parameter contributions, or relationships among objectives.
These signals characterize the local optimization condition, but do not
directly reveal what would happen if the model were actually updated
along different directions. This distinction is particularly important
in head avatar modeling, where geometry-, appearance-, and
consistency-related objectives interact through a tightly coupled
representation. A larger loss or stronger gradient does not necessarily
imply that emphasizing the corresponding direction will yield the
greatest subsequent improvement.

To address this problem, we propose a counterfactual route optimization
framework for Gaussian head avatar modeling, which evaluates alternative
short-horizon updates from the same training state to estimate the
relative utility of different optimization routes. Rather than adjusting optimization emphasis through predefined heuristics or proxy statistics, we first evaluate several short-horizon candidate optimization routes from the same training state and use their actual outcomes to obtain route-level optimization evidence. We then organize the resulting counterfactual evidence into a factorized representation that separately captures the relative preference between geometry- and appearance-oriented optimization and the additional benefit of joint optimization. Finally, we amortize this state-dependent route evidence into a lightweight route controller, which directly predicts the current optimization tendency during full training and applies bounded modulation to the corresponding training objectives. In this way, our method enables dynamic optimization coordination while preserving the underlying avatar representation and training framework.

Our main contributions are summarized as follows:
\begin{itemize}
    \item We introduce counterfactual route optimization for head avatar
training, enabling state-dependent optimization preference estimation
based on the realized outcomes of alternative updates.

    \item A factorized representation of counterfactual route evidence is
introduced to separately capture geometry--appearance preference and
the additional benefit of joint optimization, providing a richer
description of optimization demand than hard route selection.

    \item We amortize the resulting state-dependent route evidence into a lightweight controller for bounded modulation of the training objectives during full training, enabling adaptive optimization coordination with low online overhead while preserving the original avatar representation and training framework.
\end{itemize}

\section{Related Work}

\subsection{Animatable Head Avatar Representations}

Animatable head avatars have evolved from implicit neural fields to explicit Gaussian and hybrid representations. Implicit approaches model geometry and appearance in continuous neural fields, while recent Gaussian-based methods improve rendering efficiency and deformation flexibility through surface binding, learned deformation, and more expressive Gaussian parameterizations \citep{zheng2022imavatar,qian2024gaussianavatars,xiang2024flashavatar,zhang2025hravatar,feng2025gpavatar}. Hybrid formulations further combine structured facial geometry with flexible appearance primitives, as explored in MeGA, SVG-Head, and PhysHead \citep{wang2025mega,sun2025svghead,kabadayi2026physhead}. As these representations incorporate increasingly heterogeneous geometry, appearance, and deformation components, reconstruction quality depends not only on representation capacity, but also on how different optimization objectives are coordinated during training.

\subsection{Optimization Strategies for Avatar Reconstruction}

Recent avatar methods increasingly organize reconstruction through progressive or structured optimization instead of maintaining an unchanged optimization emphasis throughout training. TetGS separates geometry and appearance optimization into different stages, Fresco progressively shifts supervision from low-frequency stabilization toward high-frequency refinement, and ProgressiveAvatars gradually expands representation capacity \citep{liu2025tetgs,zhang2026fresco,song2026progressiveavatars}. Other methods introduce hierarchical refinement or structured consistency constraints to improve reconstruction stability, such as SVG-Head and MVCHead \citep{sun2025svghead,chharia2026mvchead}. These approaches demonstrate the importance of coordinating heterogeneous objectives over the course of training. However, their optimization schedules, activation patterns, or relative emphasis are largely specified in advance, rather than determined according to the optimization demand of each individual training state.

\subsection{Adaptive and Multi-Objective Optimization}

Adaptive optimization has been widely studied in multi-task and
multi-objective learning, where optimization priorities are adjusted
according to the current learning state rather than remaining fixed.
Recent methods estimate task importance from parameter contributions,
task-specific updates, bi-level objectives, or inter-task interactions
\citep{jeong2024taskpriority,hervella2024mao,qin2025consmtl,phan2025beyond},
while other studies analyze the trade-offs of dynamic objective balancing
and aligned multi-objective settings
\citep{chen2024threeway,efroni2025aligned}.

Such adaptive coordination is also relevant to head avatar reconstruction.
However, existing methods typically infer optimization preference from
instantaneous losses, gradients, parameter contributions, or
meta-objectives. These signals describe the current optimization
condition, but do not directly measure the effect of taking different
update directions from the same training state. Determining
state-dependent optimization preference from the realized outcomes of
alternative updates therefore remains underexplored in head avatar
modeling.

\begin{figure*}[t]
  \centering
  
  \includegraphics[width=\textwidth]{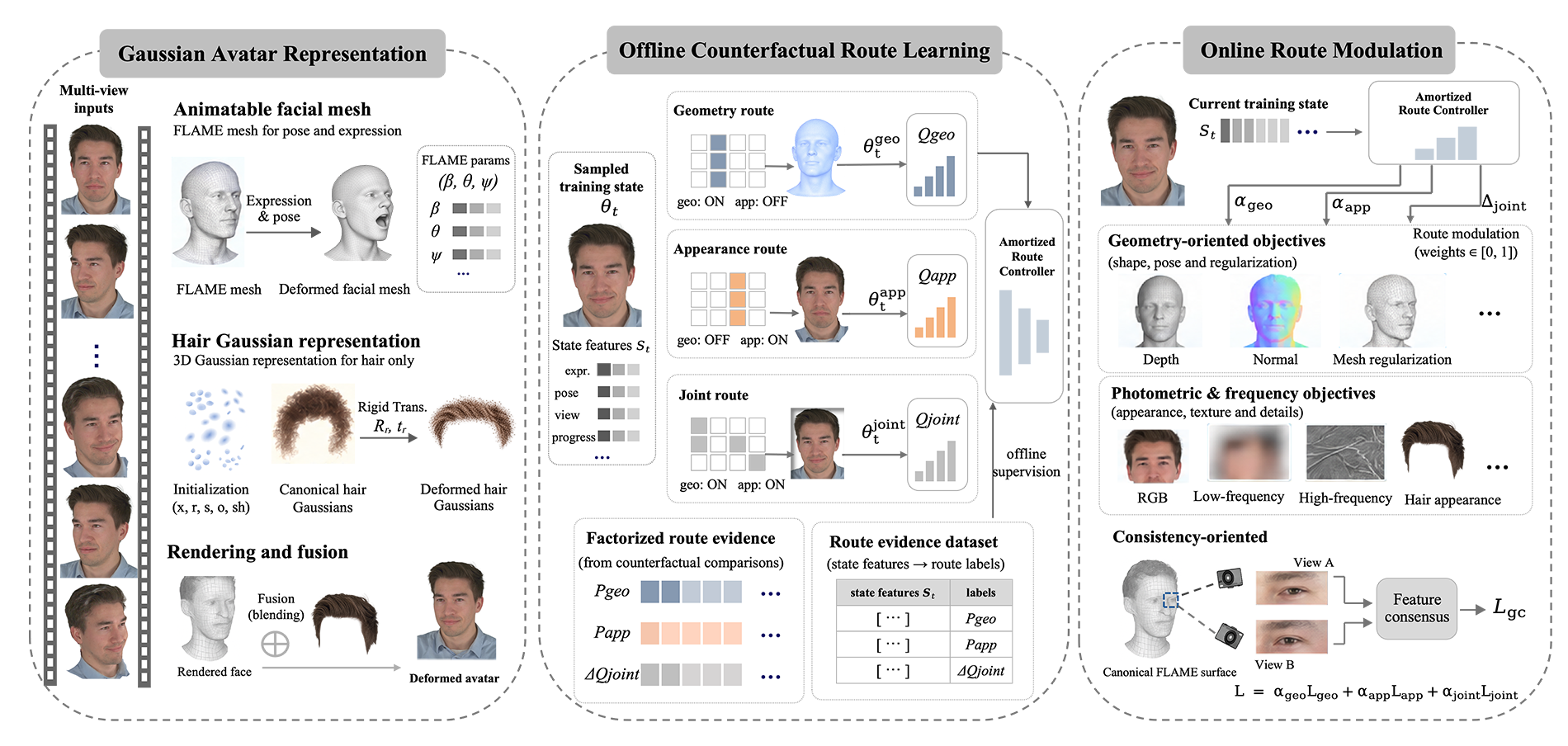}

   \caption{Overview of the proposed framework.
Multi-view observations are first integrated into a hybrid head avatar, with the face modeled by an animatable FLAME mesh and the hair represented by 3D Gaussians, which are jointly rendered to produce the final avatar.
In the \emph{offline counterfactual route learning} stage, geometry, appearance, and joint routes are instantiated from the same sampled training state and evaluated through route-restricted lookahead updates.
Their realized utilities are then factorized into geometry--appearance preference and residual joint advantage, forming route evidence that is paired with the corresponding state features to supervise an amortized route controller.
During \emph{online route modulation}, the controller infers the current optimization preference from the training state and accordingly modulates geometry-oriented, photometric and frequency, and consistency-oriented objectives during training.}
   \label{fig:network}
\end{figure*}

\section{Method}

Our method introduces state-dependent route coordination into head avatar optimization. As illustrated in Fig.~\ref{fig:network}, starting from the same training state, we evaluate short-horizon geometry-oriented, appearance-oriented, and joint update routes, convert their realized outcomes into factorized route evidence, and amortize this evidence into a lightweight controller for low-cost modulation during full training. In the following, we first introduce the underlying avatar representation and baseline optimization, and then describe the counterfactual route evaluation, factorized route evidence, and amortized route optimization.

\subsection{Avatar Representation and Baseline Optimization}

We adopt an animatable head representation~\citep{wang2025mega} that
models the facial region with an explicit parametric surface and the hair
region with 3D Gaussian primitives. At frame $t$, we represent the complete avatar as:
\begin{equation}
    \mathcal{A}_t = \{\mathcal{M}_t,\mathcal{G}_t\},
\end{equation}
where $\mathcal{M}_t$ is the posed facial mesh and $\mathcal{G}_t$ represents the corresponding hair Gaussian field. The facial geometry is parameterized by FLAME as:
\begin{equation}
    \mathcal{M}_t =
    \mathcal{M}(\boldsymbol{\beta},
    \boldsymbol{\psi}_t,
    \boldsymbol{\phi}_t),
\end{equation}
where $\boldsymbol{\beta}$ denotes the identity-dependent shape,
and $\boldsymbol{\psi}_t$ and $\boldsymbol{\phi}_t$ denote the
expression and pose at frame $t$, respectively. The hair region is
represented by a set of anisotropic Gaussian primitives:
\begin{equation}
    \mathcal{G}_t =
    \left\{
    (\boldsymbol{\mu}_i,
    \boldsymbol{\Sigma}_i,
    \mathbf{c}_i,
    \alpha_i)
    \right\}_{i=1}^{N},
\end{equation}
where $\boldsymbol{\mu}_i \in \mathbb{R}^{3}$ is the Gaussian center,
$\boldsymbol{\Sigma}_i$ is the anisotropic covariance,
$\mathbf{c}_i$ denotes the appearance attributes, and
$\alpha_i$ is the opacity.
The Gaussian field is deformed together with the current head state,
allowing the hair representation to follow the animated avatar across
different expressions and poses.

Given camera parameters $\Pi_t^v$ for view $v$, the facial surface and
hair field are rendered through mesh rasterization and Gaussian
splatting, respectively, and are subsequently composited to form the
final rendering:
\begin{equation}
    \hat{\mathbf I}_t^v =
    \mathcal{B}\!\left(
    \mathcal{R}_{\mathrm{mesh}}(\mathcal{M}_t,\Pi_t^v),
    \mathcal{R}_{\mathrm{gs}}(\mathcal{G}_t,\Pi_t^v)
    \right),
\end{equation}
where $\mathcal{R}_{\mathrm{mesh}}$ and $\mathcal{R}_{\mathrm{gs}}$
denote the differentiable renderers for the mesh and Gaussian
components, and $\mathcal{B}$ denotes their compositing operation.

Training follows the reconstruction objective and optimization schedule
of the underlying avatar framework. The supervision jointly constrains
the rendered appearance, facial geometry, and representation
regularity, while stage-dependent auxiliary terms may be activated as
training proceeds. We denote all optimizable avatar parameters by
$\Theta$ and the resulting training state at iteration $t$ by
$\Theta_t$. This state summarizes the current solution reached under
the baseline optimization and serves as the common starting point for
the alternative parameter updates considered in the following
counterfactual analysis.

\subsection{Counterfactual Route Evaluation}

Given the current optimization state $\Theta_t$, we seek to determine
which update direction is locally more effective at this point of
training. Rather than inferring such preference from pre-update
statistics, such as loss magnitude or gradient strength, we directly
compare the outcomes induced by alternative parameter updates.

We divide the optimizable parameters into two functional subsets,
$S_g$ and $S_a$, according to their dominant roles in reconstruction.
$S_g$ mainly contains parameters associated with geometric structure
and deformation, including facial shape and deformation variables,
Gaussian spatial parameters, and hair deformation parameters.
$S_a$ contains parameters primarily responsible for appearance
modeling, such as neural textures, appearance decoders, and Gaussian
color coefficients. This partition is introduced only to instantiate
controlled update directions, and we do not assume strict
geometry--appearance disentanglement. For a route $r$, the trainable
parameter subset is defined as:
\begin{equation}
S_r =
\begin{cases}
S_g, & r=g,\\
S_a, & r=a,\\
S_g \cup S_a, & r=j,
\end{cases}
\label{eq:route_subset}
\end{equation}
where $g$, $a$, and $j$ denote the geometry, appearance, and joint
routes, respectively. Parameters outside $S_r$ are held fixed during
the corresponding update.

We evaluate each route through a lookahead operation. Starting from the
same state $\Theta_t$, a lookahead temporarily applies a small number of
optimization steps to the parameter subset $S_r$ using the same
mini-batch $B_t$ and training objective $L_t$:
\begin{equation}
\Theta_t^r
=
U^{(m)}(\Theta_t, B_t, L_t; S_r),
\label{eq:route_update}
\end{equation}
where $U^{(m)}$ denotes an $m$-step route-restricted update and
$\Theta_t^r$ is the temporary state obtained under route $r$. In our
implementation, we use a single-step lookahead, i.e., $m=1$. The
resulting state is used only for route evaluation and is not retained in
the main optimization trajectory.

This construction provides the counterfactual comparison used in our
method. From the same training state $\Theta_t$, we separately apply
geometry-, appearance-, and joint-oriented updates and examine the
outcomes that each update would produce. Since these alternative
updates are evaluated from an identical starting point but are not
simultaneously committed to the actual training trajectory, their
resulting states provide counterfactual evidence about the local effect
of different optimization directions.

To assess these outcomes, we consider the individual loss terms used by
the current training objective and denote their set by
$\mathcal{K}_t$. Depending on the training stage,
$\mathcal{K}_t$ may include photometric, geometric, regularization,
frequency, and consistency terms. Let $l_k$ denote the $k$-th loss term
in $\mathcal{K}_t$. Its relative improvement under route $r$ is defined as:
\begin{equation}
\Delta_k^r
=
\frac{
l_k(\Theta_t)-l_k(\Theta_t^r)
}{
|l_k(\Theta_t)|+\epsilon
},
\qquad k\in\mathcal{K}_t,
\label{eq:route_improvement}
\end{equation}
where $\epsilon$ is a small constant for numerical stability. A positive
$\Delta_k^r$ indicates an improvement in the corresponding loss term.
Since different loss terms can have substantially different scales and
response ranges, their relative improvements are normalized
independently before aggregation.

Reducing some loss terms may still be accompanied by degradation of
geometric structure. We therefore define a subset
$\mathcal{H}_t \subseteq \mathcal{K}_t$ containing the geometric terms
that directly reflect structural quality, including depth, normal, and
mesh regularization terms. For each $k\in\mathcal{H}_t$, we measure the adverse change as: 
\begin{equation}
D_k^r
=
\max\left(
0,
\frac{
l_k(\Theta_t^r)-l_k(\Theta_t)
}{
|l_k(\Theta_t)|+\epsilon
}
\right),
\qquad k\in\mathcal{H}_t.
\label{eq:route_degradation}
\end{equation}
Thus, $D_k^r$ is zero when the corresponding geometric term is preserved
or improved, and becomes positive only when the route causes structural
degradation. After component-wise normalization, the utility of route $r$ is defined as:
\begin{equation}
Q_r
=
\sum_{k\in\mathcal{K}_t}
w_k\,\widetilde{\Delta}_k^r
-
\lambda_h
\sum_{k\in\mathcal{H}_t}
w_k\,\widetilde{D}_k^r,
\label{eq:route_utility}
\end{equation}
where $\widetilde{\Delta}_k^r$ and
$\widetilde{D}_k^r$ denote the normalized improvement and degradation,
$w_k$ is the weight of the corresponding loss term, and $\lambda_h$
controls the penalty on structural degradation. The first term rewards routes that yield improvements across the current reconstruction objectives, while the second penalizes routes whose gains are accompanied by deterioration in geometry-sensitive terms. Thus, $Q_r$ provides a balanced measure for comparing the realized local effects of different routes at the current training state.

Applying the above evaluation to the three routes yields
$Q_g$, $Q_a$, and $Q_j$, which provide state-dependent evidence grounded
in the realized outcomes of alternative updates. These utilities play
different roles in characterizing the current optimization demand.
While $Q_g$ and $Q_a$ describe the relative preference between
geometry- and appearance-oriented updates, $Q_j$ provides the reference
for assessing whether coordinated optimization offers additional
benefit beyond either individual route. This asymmetric interpretation
motivates the factorized route evidence introduced in the next section.

\subsection{Factorized Route Evidence}

A direct three-way interpretation of $Q_g$, $Q_a$, and $Q_j$ would obscure an important asymmetry among the candidate routes. The geometry and appearance routes represent two alternative directional updates, whereas the joint route serves a different role: it reveals whether coordinating both parameter subsets provides value beyond committing to either individual direction. Treating all three routes as equivalent alternatives would therefore entangle directional preference with collaborative benefit. To preserve this distinction, we factorize the counterfactual evidence into a geometry--appearance preference and a residual joint advantage.

We first convert the geometry and appearance utilities into a pair of normalized preference scores, $P_g$ and $P_a$, representing the relative preference for geometry- and appearance-oriented updates, respectively:
\begin{equation}
P_g =
\frac{\exp(Q_g/\tau)}
{\exp(Q_g/\tau)+\exp(Q_a/\tau)},
\qquad
P_a = 1-P_g,
\label{eq:geo_app_preference}
\end{equation}
where $\tau$ controls the sharpness of the preference distribution. Rather than reducing the comparison to a hard route selection, $(P_g,P_a)$ provides a continuous characterization of the relative optimization preference between the two update directions. We then isolate the benefit that cannot be explained by either individual route alone. Specifically, we define the residual joint advantage as:
\begin{equation}
\Delta Q_j =
Q_j-\max(Q_g,Q_a).
\label{eq:joint_advantage}
\end{equation}
By referencing the stronger individual route, $\Delta Q_j$ measures the additional benefit of coordinated optimization rather than the absolute quality of the joint update. A positive value indicates that jointly adapting the two parameter subsets provides additional utility beyond the better individual route, whereas a non-positive value indicates no such advantage at the current state. We denote the geometry--appearance preference by:
\begin{equation}
\mathbf{P}_t^{ga} = (P_g,P_a),
\end{equation}
and summarize the factorized route evidence at state $\Theta_t$ as:
\begin{equation}
E_t =
\left(
\mathbf{P}_t^{ga},
\Delta Q_j
\right).
\label{eq:factorized_evidence}
\end{equation}
This representation explicitly separates directional preference from collaborative gain, providing a compact and interpretable characterization of the optimization demand at the current state. More importantly, it avoids forcing qualitatively different route effects into a single categorical decision, and provides the supervision through which counterfactual route evidence is amortized into the online optimization process.

\subsection{Amortized Route Optimization}

Although counterfactual evaluation provides informative state-dependent
route evidence, repeatedly executing multiple route-restricted updates
during full training would introduce substantial computational overhead.
We therefore amortize the offline evidence into a lightweight controller
$f_{\phi}$:
\begin{equation}
(\hat{P}_g,\hat{P}_a,\widehat{\Delta Q}_j)
=
f_{\phi}(\mathbf{x}_t),
\label{eq:route_controller}
\end{equation}
where $\mathbf{x}_t$ denotes a compact descriptor of the current
optimization state. The controller is supervised by the factorized
evidence defined in Sec.~3.3, replacing explicit counterfactual
evaluation during full training with a single forward prediction.

The offline and online stages use route evidence differently.
During counterfactual evaluation, each route is isolated by updating
only its associated parameter subset while keeping the remaining
parameters fixed. During online training, parameter updates are no
longer restricted. Instead, the predicted route evidence modulates the
corresponding training objectives within a bounded range.

To realize this modulation, we organize the training objectives according
to their dominant optimization effects into geometry-oriented,
appearance-oriented, and consistency-oriented groups. This grouping is
not intended to imply mutually exclusive effects; instead, it provides a
low-dimensional optimization abstraction for regulating the relative
emphasis among different update directions. The detailed objective
assignment is provided in the supplementary material.

The factorized route evidence induces two complementary levels of
control. The geometry--appearance preference first determines the
relative emphasis between the two update directions:
\begin{equation}
s_g =
\operatorname{clip}
\left(
1+\gamma(\hat{P}_g-\hat{P}_a),
s_{\min},s_{\max}
\right),
\qquad
s_a =
\operatorname{clip}
\left(
1+\gamma(\hat{P}_a-\hat{P}_g),
s_{\min},s_{\max}
\right),
\label{eq:ga_scale}
\end{equation}
while the residual joint advantage provides an additional coordinated
amplification:
\begin{equation}
s_j =
\operatorname{clip}
\left(
1+\gamma\max(0,\widehat{\Delta Q}_j),
s_{\min},s_{\max}
\right).
\label{eq:joint_scale}
\end{equation}
Here, $\gamma$ controls the modulation strength, and $s_{\min}$ and
$s_{\max}$ denote the lower and upper bounds of the modulation factors,
respectively. The first level determines the relative preference between
geometry- and appearance-oriented updates, whereas the second determines
whether both directions should receive additional coordinated
amplification.

Let $L_g$, $L_a$, and $L_c$ denote the geometry-oriented,
appearance-oriented, and consistency-oriented objective groups,
respectively. The final training objective is written as:
\begin{equation}
L_{\mathrm{mod}}
=
L_{\mathrm{base}}
+
(s_g s_j-1)L_g
+
(s_a s_j-1)L_a
+
(s_j-1)L_c,
\label{eq:modulated_loss}
\end{equation}
where $L_{\mathrm{base}}$ denotes the original training objective. This
formulation treats the baseline optimization as the reference and
applies only bounded state-dependent corrections. The
geometry--appearance preference controls the relative balance between
the two directions, while the joint factor modulates their shared
optimization strength.

For the consistency-oriented objective, we employ group-level cross-view
consistency to associate corresponding local observations through the
canonical facial structure and encourage consistent feature responses
across synchronized views. Together with the directional and coordinated
modulation above, the offline counterfactual evidence is thereby
translated into efficient state-dependent optimization without requiring
online route rollouts during full training.

\begin{figure*}[t]
  \centering
  
  \includegraphics[width=0.95\textwidth]{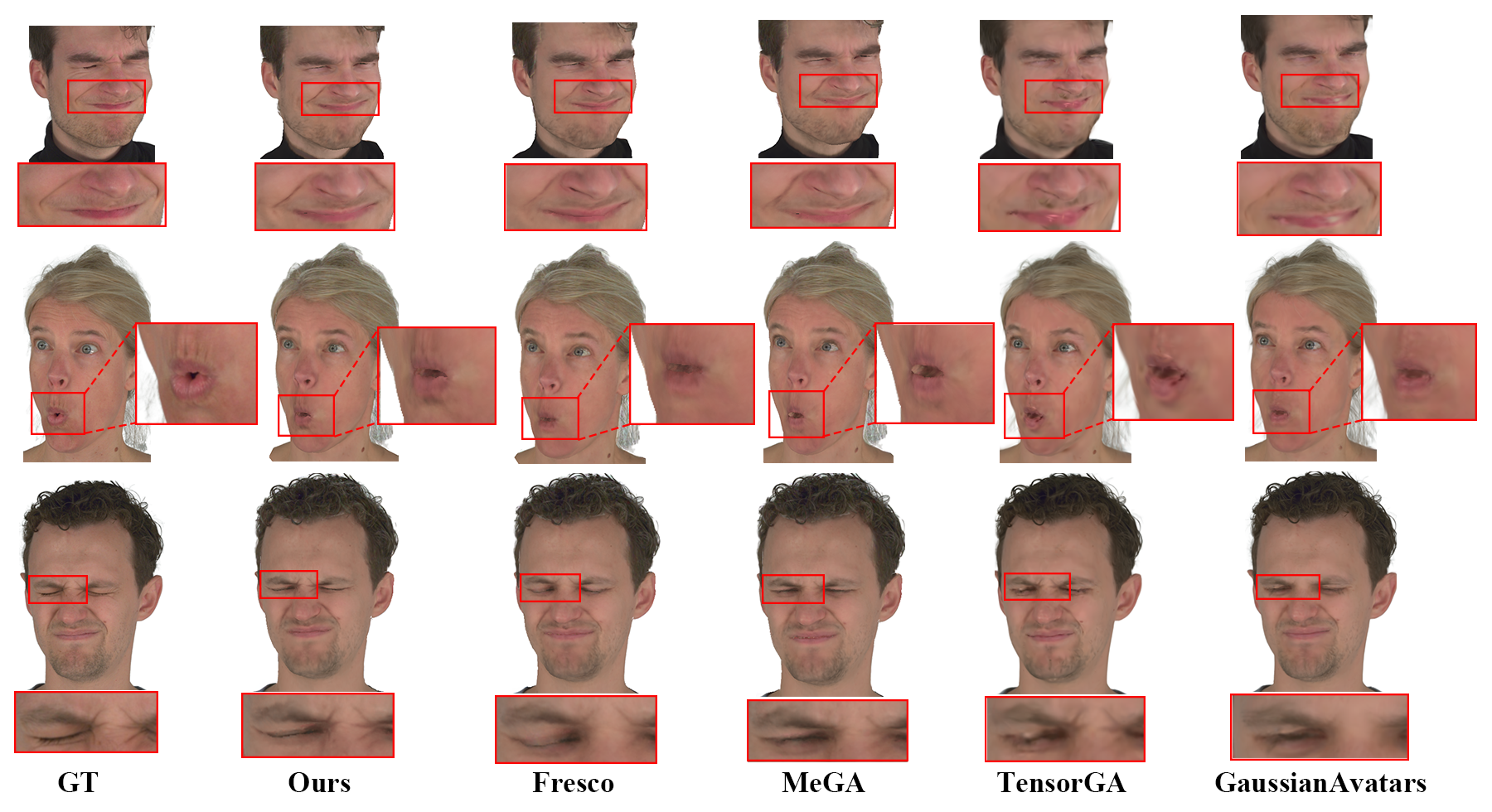}

   \caption{Qualitative comparison on self-reenactment of head avatars. Our results preserve more accurate mouth shapes and eye structures under expression changes.}
   \label{fig:self_reenactment}
\end{figure*}

\section{Experiments}

\subsection{Experimental Setup}

\textbf{Dataset.}
We conduct experiments on the NeRSemble dataset, which provides
synchronized multi-view recordings of multiple subjects captured in a
controlled studio using 16 calibrated cameras. Following the evaluation protocols adopted by GaussianAvatars~\citep{qian2024gaussianavatars}
and MeGA~\citep{wang2025mega}, we select 8 subjects for evaluation.
Each subject contains eleven sequences, including ten sequences with
predefined expressions or emotions and one free-form sequence.
All frames are resized to $802\times550$, with foreground masks
generated for each view. We additionally obtain semantic facial parsing
maps and reconstruct per-frame depth maps using Metashape.
For each subject, 9 predefined sequences and 15 camera views are used
for training, while one predefined sequence and the remaining camera
view are reserved for evaluation.

\begin{table}[t]
\centering
\caption{Quantitative comparison on novel-view synthesis and
self-reenactment. Results are averaged over 8 subjects.}
\label{tab:main_comparison}
\small
\setlength{\tabcolsep}{3.8pt}
\renewcommand{\arraystretch}{1.08}
\begin{tabular}{l|ccc|ccc}
\toprule
\multirow{2}{*}{Method}
& \multicolumn{3}{c|}{Novel-View}
& \multicolumn{3}{c}{Self-Reenactment} \\
& PSNR$\uparrow$ & SSIM$\uparrow$ & LPIPS$\downarrow$
& PSNR$\uparrow$ & SSIM$\uparrow$ & LPIPS$\downarrow$ \\
\midrule
Gaussian Head Avatar ~\citep{xu2024gaussian}
& 29.84 & 0.895 & 0.083
& 22.34 & 0.851 & 0.145 \\

EAvatar ~\citep{zhang2025eavatar}
& 30.16 & 0.898 & 0.082
& 22.56 & 0.852 & 0.145 \\

GaussianAvatars ~\citep{qian2024gaussianavatars}
& 31.50 & 0.935 & 0.060
& 29.75 & 0.930 & 0.066 \\

MeGA ~\citep{wang2025mega}
& 30.79 & 0.932 & 0.066
& 29.62 & 0.926 & 0.071 \\

TensorGA ~\citep{wang20253d}
& 31.28 & 0.931 & 0.067
& 29.90 & \underline{0.931} & 0.071 \\

Fresco ~\citep{zhang2026fresco}
& \underline{31.80} & \underline{0.938} & \underline{0.058}
& \underline{30.30} & 0.929 & \underline{0.064} \\

Ours
& \textbf{31.90} & \textbf{0.939} & \underline{0.058}
& \textbf{30.49} & \underline{0.931} & \textbf{0.063} \\
\bottomrule
\end{tabular}
\end{table}

\textbf{Settings.}
We evaluate the reconstructed avatars under novel-view synthesis,
self-reenactment, and cross-reenactment. For novel-view synthesis, we
render the avatar from the camera view excluded during training and
compare the results with the corresponding ground-truth images. For
self-reenactment, the avatar is driven by poses and expressions from the
held-out sequence of the same subject and rendered across all 16 camera
views. For cross-reenactment, the target avatar is driven using poses
and expressions extracted from another subject, and the results are
evaluated qualitatively due to the absence of paired ground truth.
For quantitative evaluation, we report PSNR, SSIM, and LPIPS, with all
metrics computed within the same foreground region to exclude the
influence of background pixels.

\begin{figure*}
  \centering
  
  \includegraphics[width=0.9\textwidth]{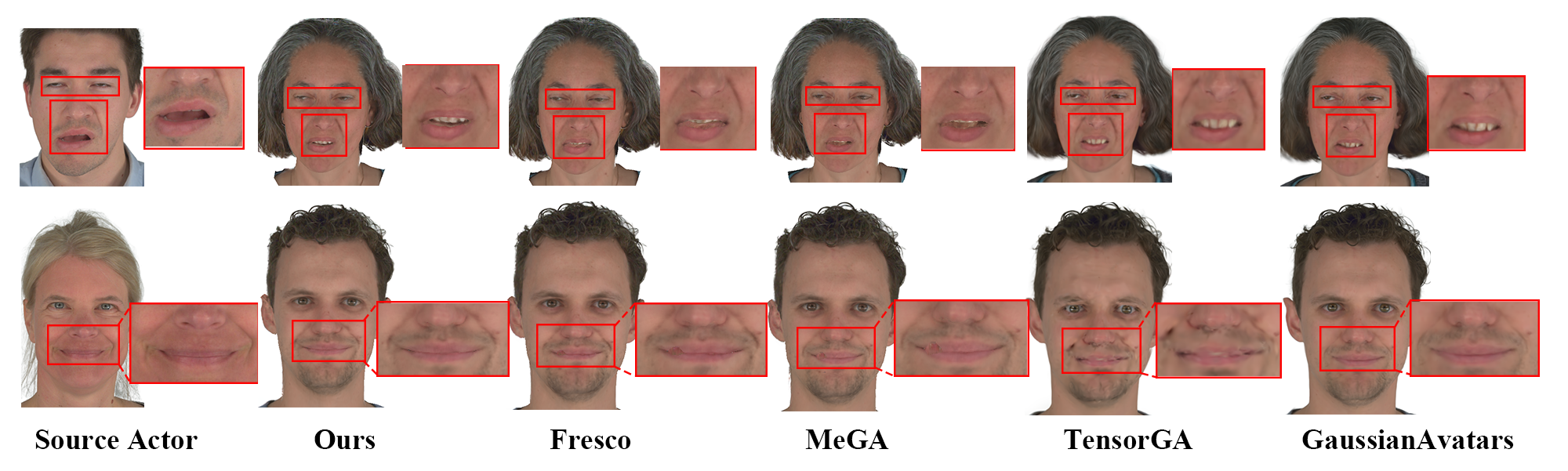}

   \caption{Qualitative comparison on cross-reenactment of head avatars. Our results
show cleaner lip boundaries and more coherent mouth structures under cross-identity expression transfer.}
   \label{fig:cross_reenactment}
\end{figure*}
\subsection{Comparisons with State-of-the-Art Methods}

We compare our method with representative state-of-the-art animatable
head-avatar approaches, including Gaussian Head Avatar~\citep{xu2024gaussian},
EAvatar~\citep{zhang2025eavatar}, GaussianAvatars~\citep{qian2024gaussianavatars},
MeGA~\citep{wang2025mega}, TensorGA~\citep{wang20253d}, and
Fresco~\citep{zhang2026fresco}. These methods cover implicit,
Gaussian-based, and hybrid avatar representations with different
optimization strategies. All methods are evaluated under the same data
split and evaluation protocol described above for fair comparison. 
As shown in Table~\ref{tab:main_comparison}, our method achieves
consistent performance advantages across both novel-view synthesis and
self-reenactment. Compared with Fresco, the two methods achieve relatively
close quantitative performance, while our method shows more consistent
advantages across the two evaluation settings. The subsequent qualitative
comparisons further show that these improvements are also reflected in
the preservation of local facial structures and fine details. 
Figures~\ref{fig:self_reenactment} and~\ref{fig:cross_reenactment} show the
qualitative comparisons on self-reenactment and cross-reenactment.
Our method preserves clearer local structures and finer details,
especially around the mouth, teeth, and eye regions, while competing
methods more often exhibit blurred boundaries or incomplete local
structures. Additional qualitative comparisons on novel-view synthesis
and local reconstruction error visualizations are provided in the
supplementary material (Figs.~\ref{fig:novel_view} and~\ref{fig:error_vis}).

\subsection{Ablation Study}
\paragraph{Effect of Route Modulation.}
We first analyze the contribution of the proposed route modulation
strategy to the final reconstruction quality on subject 264. Starting from the baseline, we introduce fixed-weight group consistency (Fixed-GC) to isolate the
effect of additional cross-view supervision. We further construct a
static \emph{Mean Mod.} variant by averaging the modulation factors
produced by the full controller over training and using these averages
as fixed scaling factors in a separate retraining run. This provides a
direct control for examining whether the improvement of the full model
can be attributed simply to a better fixed weighting configuration.
Detailed weight settings and training configurations are provided in the
supplementary material.

As shown in Table~\ref{tab:ablation_overall}, Fixed-GC improves over the
baseline, confirming that cross-view consistency itself is beneficial.
Mean Mod. further improves the overall reconstruction quality, indicating
that the modulation learned by the controller also captures a more
favorable global weighting than the original fixed configuration.
However, the full model still achieves the strongest performance.
Since Mean Mod. uses the average modulation of the full controller but
removes its temporal variation, the remaining gap shows that a better
fixed weighting alone is insufficient to account for the full gain.
The state-dependent adjustment of optimization emphasis therefore
provides an additional contribution during training.

\begin{figure*}[t]
  \centering
  
  \includegraphics[width=0.9\textwidth]{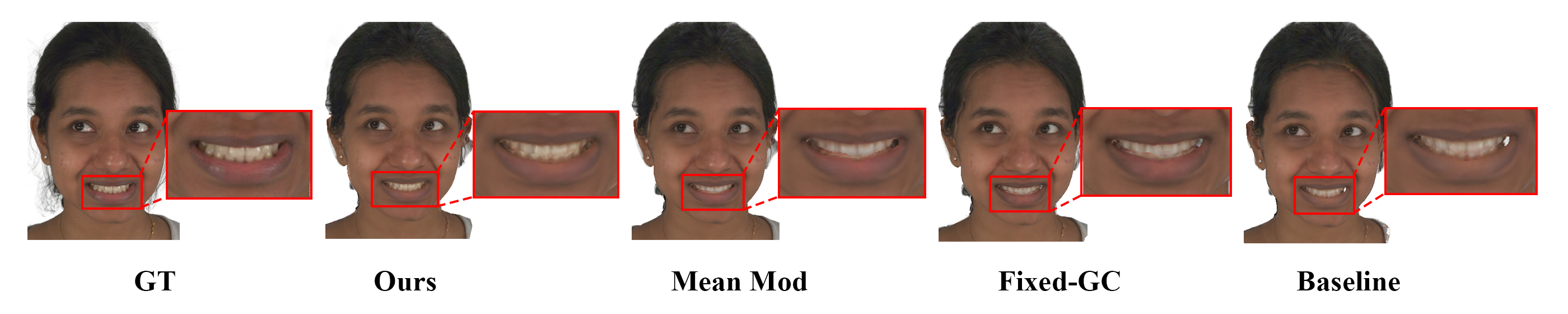}

   \caption{Qualitative ablation of the route modulation strategy. The enlarged mouth regions show progressively reduced artifacts, with the full model recovering more complete teeth and cleaner lip boundaries.}
   \label{fig:ablation_overall}
\end{figure*}

The qualitative results in Fig.~\ref{fig:ablation_overall} exhibit a
similar progression. The Baseline still contains noticeable local
artifacts around the mouth region. These artifacts are reduced in
Fixed-GC and further alleviated by Mean Mod. The full model recovers
more complete tooth structures and cleaner lip boundaries, with local
details that are closer to the ground truth.

\begin{table}[t]
\centering
\caption{Ablation study of the route modulation strategy on novel-view synthesis and self-reenactment.}
\label{tab:ablation_overall}
\small
\setlength{\tabcolsep}{3.8pt}
\renewcommand{\arraystretch}{1.08}
\begin{tabular}{l|ccc|ccc}
\toprule
\multirow{2}{*}{Method}
& \multicolumn{3}{c|}{Novel-View}
& \multicolumn{3}{c}{Self-Reenactment} \\
& PSNR$\uparrow$ & SSIM$\uparrow$ & LPIPS$\downarrow$
& PSNR$\uparrow$ & SSIM$\uparrow$ & LPIPS$\downarrow$ \\
\midrule
Baseline
& 30.36 & 0.952 & 0.066
& 29.60 & 0.948 & 0.067 \\

Fixed-GC
& 30.98 & \underline{0.954} & \underline{0.060}
& 30.04 & 0.950 & 0.064 \\

Mean Mod.
& \underline{31.33} & \underline{0.954} & 0.061
& \underline{30.19} & \underline{0.952} & \underline{0.062} \\

Ours
& \textbf{31.71} & \textbf{0.956} & \textbf{0.057}
& \textbf{30.98} & \textbf{0.954} & \textbf{0.060} \\
\bottomrule
\end{tabular}
\end{table}

\section{Conclusion}

This work approaches head avatar reconstruction from an optimization
perspective, focusing on how different update directions should be
coordinated as training evolves. Instead of relying on a fixed
optimization configuration, we use the realized effects of alternative
updates to characterize the current optimization preference and further
translate this information into efficient state-dependent modulation.
In this way, the proposed framework adapts the optimization emphasis
during training while preserving the underlying avatar representation
and reconstruction pipeline. Experiments on the NeRSemble dataset
demonstrate improved reconstruction quality on novel-view synthesis and
reenactment tasks, showing the effectiveness of state-dependent route
coordination for high-fidelity head avatar modeling.

\subsection*{AI use statement}

In this work, we used generative AI tools primarily for language editing
and grammatical refinement of the manuscript. We also used generative AI
tools to assist with reference verification, provide limited feedback on
research methodology and experimental design, and conduct preliminary
manuscript review and readability checks. All AI-assisted edits and
suggestions were manually reviewed by the authors, and all references
were verified against their corresponding official publication records.
We take responsibility for the final content of this work, including
text, claims, and artifacts produced with the aid of generative AI.

\subsection*{Reproducibility statement}

We provide detailed implementation and training configurations in the
supplementary material, including the counterfactual route evaluation,
route-evidence construction, controller training, and online modulation
settings. The data preprocessing procedure, evaluation protocol,
ablation settings, and computational cost are also documented to
facilitate reproduction of the reported results.

\subsection*{Ethics Statement}

This work uses publicly available human face data from the NeRSemble
dataset and does not involve additional collection of personal data.
All experiments are conducted for research purposes following the
dataset's established usage protocol.

\bibliography{iclr2027_conference}
\bibliographystyle{iclr2027_conference}
\clearpage
\appendix

\section{Implementation Details}

\textbf{Training Configuration.}
All experiments are run on a single NVIDIA RTX A6000 GPU with 48\,GB
of memory. MeGA, Fresco, and our method are trained for 100 epochs per
subject using a batch size of 16 and an initial learning rate of
$1\times10^{-3}$. GaussianAvatars and TensorGA are optimized for 600K
iterations according to their official training protocols.

\textbf{Ablation Settings.}
The \emph{Baseline} follows the original MeGA reconstruction objectives
and their corresponding loss weights. For Fixed-GC, the consistency
weight is empirically set to $0.002$. We use synchronized two-view
supervision with a relative view-angle range of $25^{\circ}$--$55^{\circ}$
and a fallback target of $40^{\circ}$. Each sample uses 512 candidate
anchors and 16 valid local groups with a $41\times41$ patch size.
The minimum foreground coverage is set to $0.8$, the normal-visibility
threshold to $0.2$, and the two GT views are equally weighted when
forming the consensus target. 

\begin{figure*}[t]
  \centering
  
  \includegraphics[width=\textwidth]{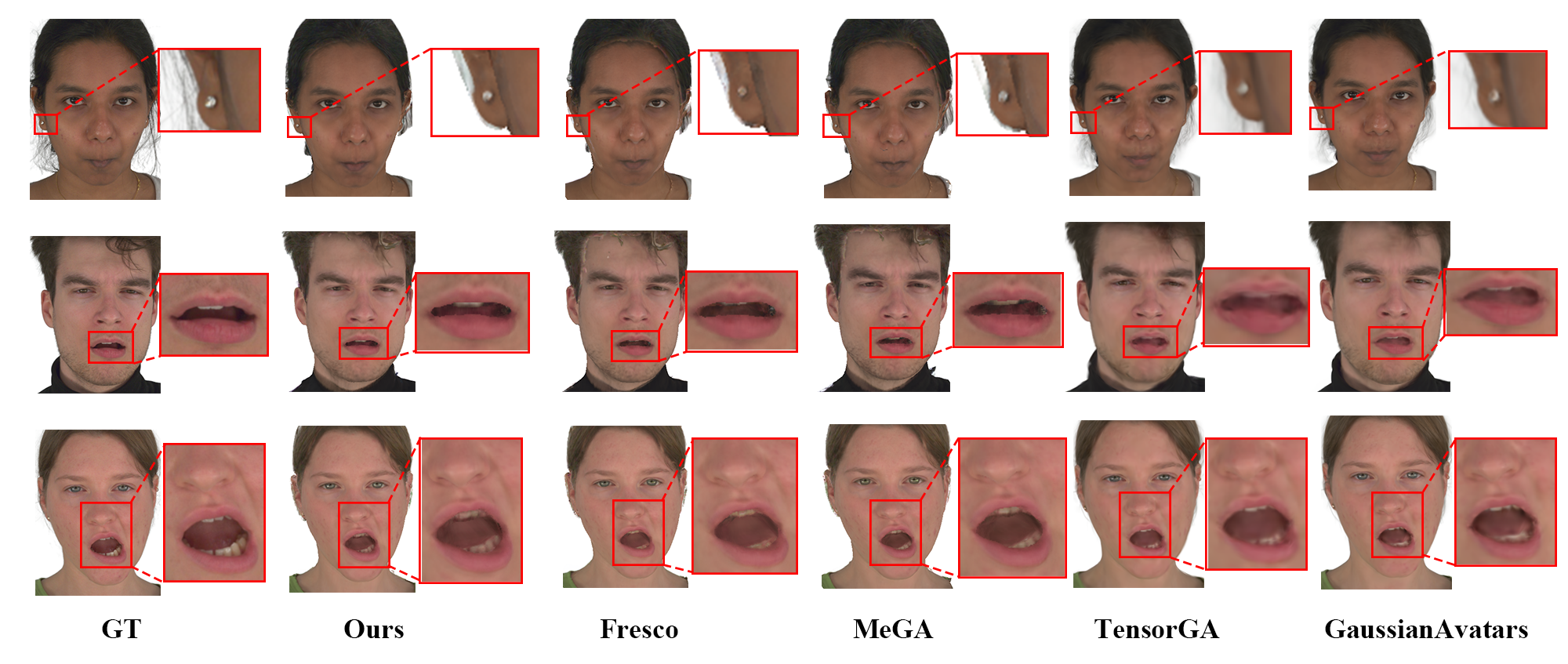}

   \caption{Qualitative comparison on novel-view synthesis of head avatars. Our results preserve cleaner local boundaries and more complete mouth and tooth structures across viewpoints, with fewer local artifacts than competing methods.}
   \label{fig:novel_view}
\end{figure*}

\textbf{Counterfactual Evidence and Route Controller.}
Offline route labels are generated from checkpoints sampled within the
first 10\% of the training trajectory. At each sampled state, geometry,
appearance, and joint routes share the same training batch and perform a
single lookahead update. Route improvements are normalized
component-wise using percentile normalization, with
$\epsilon=10^{-8}$, $\lambda_h=1$, and a softmax temperature of
$\tau=1$. The residual joint target is defined as
$Q_j-\max(Q_g,Q_a)$.

The resulting state--label pairs are used to train a lightweight route
controller with a 30-dimensional input descriptor and a hidden dimension
of 64. The controller is fixed during subsequent avatar optimization.
Online modulation starts after 20K iterations, with modulation strength
set to $0.1$ and all scaling factors clipped to $[0.9,1.1]$. The
controller is evaluated once per training iteration without additional
online lookahead updates.

\section{Computational Cost}

We report the training and inference efficiency of different head avatar
methods in Table~\ref{tab:efficiency}. All timings are measured under the
same hardware setting using a single NVIDIA RTX A6000 GPU.

\begin{table}[t]
\centering
\caption{Training and inference efficiency of different head avatar methods.}
\label{tab:efficiency}
\small
\setlength{\tabcolsep}{5pt}
\renewcommand{\arraystretch}{1.08}
\begin{tabular}{lcc}
\toprule
Method & Training Time (h)$\downarrow$ & Inference Time (ms)$\downarrow$ \\
\midrule
Gaussian Head Avatar & 45 & 79 \\
GaussianAvatars       & 7  & \textbf{17} \\
TensorGA              & 10 & 21 \\
MeGA                  & 14 & 55 \\
Fresco                & 17 & 55 \\
Ours                  & 17 & 57 \\
\bottomrule
\end{tabular}
\end{table}

\textbf{Additional Offline Cost.}
The counterfactual evidence is generated from training states sampled
within the first 10\% of the training trajectory. The complete offline
label-generation process takes approximately 1\,h 46\,min per subject,
while training the lightweight route controller requires only about
3 minutes. Both stages are performed once before full avatar
optimization.

Despite the additional training-time optimization, our method preserves
nearly the same inference efficiency as MeGA and Fresco, requiring
57\,ms per frame compared with 55\,ms for both methods. This is because
the route controller is only used during training and does not alter the
avatar rendering pipeline at inference time.

\section{Additional Results and Analysis}
We also provide additional video results covering novel-view synthesis,
self-reenactment, and cross-reenactment in the supplementary files.

\textbf{Novel-View Synthesis.}
Figure~\ref{fig:novel_view} presents additional qualitative comparisons
on novel-view synthesis. Our method preserves clearer local structures
and finer facial details across unseen viewpoints, particularly around
the mouth and teeth regions. Compared with competing methods, it
produces sharper local boundaries and more complete facial structures
with fewer view-dependent artifacts.

\begin{figure*}
  \centering
  
  \includegraphics[width=0.9\textwidth]{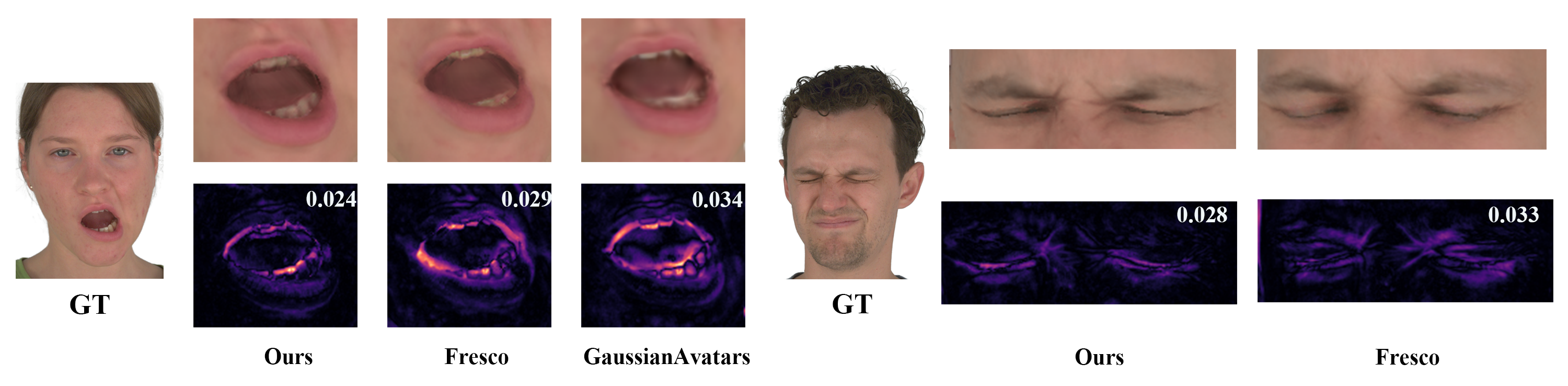}

   \caption{Local reconstruction error visualization on novel-view synthesis and
self-reenactment. The left example shows the mouth region for novel-view
synthesis, while the right example shows the eye region for
self-reenactment. Lower MAE values indicate smaller reconstruction errors.}
   \label{fig:error_vis}
\end{figure*}

\paragraph{Local Error Visualization.}
To better reveal fine-grained differences that are difficult to observe
directly from RGB comparisons, we additionally visualize local
reconstruction errors for one representative novel-view synthesis example
and one representative self-reenactment example in
Fig.~\ref{fig:error_vis}. The left example focuses on the mouth region
under novel-view synthesis, while the right example focuses on the eye
region under self-reenactment. For each displayed region, we compute the
mean absolute RGB error (MAE) between the rendered result and the
corresponding ground truth, where a lower value indicates a closer local
reconstruction. The error maps further make the spatial distribution of
the local reconstruction errors explicit. Compared with the displayed
competing methods, our method shows lower local errors in both examples,
particularly around the lip boundaries, inner-mouth structures, and eye
contours.

\textbf{Route-Evidence Ablation.}
We conduct the route-evidence ablation on subject 460 to analyze the two
signals used by the route controller. The geometry--appearance signal
determines whether the current training state should place relatively
more emphasis on geometry or appearance optimization, whereas the joint
signal determines whether both directions should be strengthened together.

Accordingly, \emph{GA-only} retains only the geometry--appearance
preference and disables joint amplification. In \emph{Joint-only}, the
geometry and appearance scales are both fixed to their neutral value,
while the controller predicts only the joint scale, which modulates the
two optimization directions together. We additionally construct a
\emph{Parallel-Route} variant that directly treats the geometry,
appearance, and joint route utilities as three parallel preferences.
The full model combines the geometry--appearance preference with the
joint signal using the proposed factorized formulation.

As shown in Table~\ref{tab:ablation_evidence}, using either signal alone
results in lower overall performance, indicating that the
geometry--appearance preference and joint modulation provide
complementary information. The Parallel-Route variant further improves
over the two single-signal variants, showing the benefit of retaining
all three route utilities. However, treating the joint route as a parallel preference still underperforms our proposed factorized formulation, since it mixes
directional preference with the additional benefit of coordinated
optimization. The full model therefore achieves
the strongest overall reconstruction quality, with the best PSNR and
LPIPS under both evaluation settings while maintaining comparable SSIM.

\begin{figure*}[t]
  \centering
  
  \includegraphics[width=\textwidth]{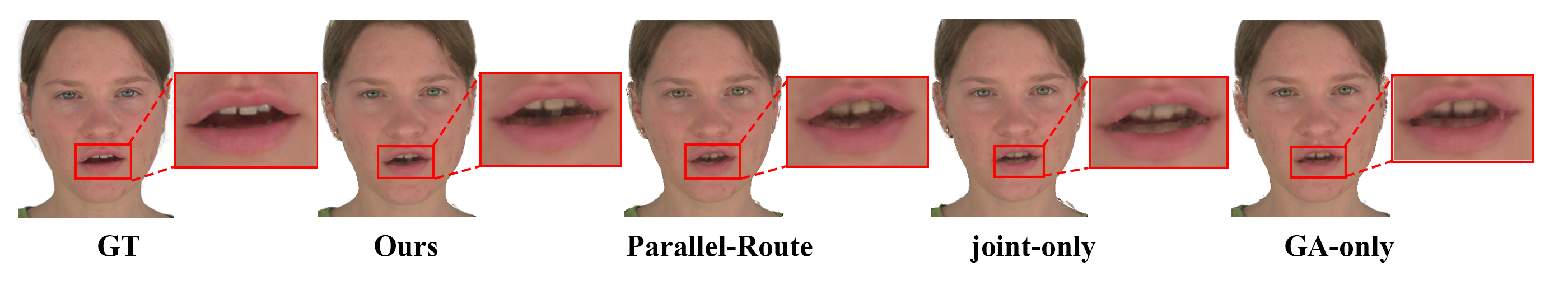}

   \caption{Qualitative comparison of the route-evidence ablation on self-reenactment.
The full model alleviates artifacts around the lower teeth and produces
a cleaner local mouth structure than the ablation variants.}
   \label{fig:ablation_evidence}
\end{figure*}

As shown in Fig.~\ref{fig:ablation_evidence}, the ablation variants
exhibit more pronounced artifacts around the lower teeth under
self-reenactment, while the full model alleviates these artifacts and
produces a cleaner local mouth structure.

\begin{table}[t]
\centering
\caption{Ablation study of the route-evidence formulation on novel-view synthesis and self-reenactment.}
\label{tab:ablation_evidence}
\small
\setlength{\tabcolsep}{4.2pt}
\renewcommand{\arraystretch}{1.08}
\begin{tabular}{l|ccc|ccc}
\toprule
\multirow{2}{*}{Variant}
& \multicolumn{3}{c|}{Novel-View}
& \multicolumn{3}{c}{Self-Reenactment} \\
& PSNR$\uparrow$ & SSIM$\uparrow$ & LPIPS$\downarrow$
& PSNR$\uparrow$ & SSIM$\uparrow$ & LPIPS$\downarrow$ \\
\midrule
GA-only
& 34.13 & 0.966 & 0.030
& 32.80 & 0.959 & 0.038 \\

Joint-only
& 33.89 & 0.966 & 0.030
& 32.84 & 0.960 & 0.038 \\

Parallel-Route
& \underline{34.78} & 0.967 & \underline{0.027}
& \underline{32.92} & \underline{0.960} & \underline{0.036} \\

Ours
& \textbf{35.38} & \textbf{0.967} & \textbf{0.025}
& \textbf{33.14} & \textbf{0.961} & \textbf{0.035} \\
\bottomrule
\end{tabular}
\end{table}

\end{document}